\documentclass[conference]{IEEEtran}

\usepackage{cite}
\usepackage{amsmath,amssymb,amsfonts}
\usepackage{bbm}
\usepackage{algorithmic}
\usepackage{graphicx}
\usepackage{textcomp}
\usepackage{xcolor}
\usepackage{booktabs}
\usepackage{multirow}
\usepackage{float}
\usepackage{algorithm}
\usepackage{tabularx}
\usepackage{adjustbox}
\usepackage{makecell}
\usepackage{svg}
\usepackage{indentfirst}

\usepackage{bm}
\usepackage[colorlinks=true, allcolors=blue]{hyperref}
\usepackage{microtype}

\def\BibTeX{{\rm B\kern-.05em{\sc i\kern-.025em b}\kern-.08em
    T\kern-.1667em\lower.7ex\hbox{E}\kern-.125emX}}
\begin{document}

\newcommand{\TableOneStyle}{%
  \small
  \renewcommand{\arraystretch}{1.15}
  \setlength{\tabcolsep}{4.5pt}
}

\title{Seeing the Needle in the Haystack: Towards Weakly-Supervised Log Instance Anomaly Localization via Counterfactual Perturbation}

\author{
\IEEEauthorblockN{Yutszyuk~Wong}
\IEEEauthorblockA{\textit{Jinan University} \\
Guangzhou, China \\
tszyuk1207@gmail.com}

\and
\IEEEauthorblockN{Wentai~Wu\textsuperscript{*}}
\IEEEauthorblockA{\textit{Jinan University} \\
Guangzhou, China \\
wentaiwu@jnu.edu.cn}

\and
\IEEEauthorblockN{Yuen-Ying Yeung}
\IEEEauthorblockA{\textit{Jinan University} \\
Guangzhou, China \\
yuengwinniee@gmail.com}

\and
\IEEEauthorblockN{Weiwei~Lin}
\IEEEauthorblockA{\textit{South China University of Technology} \\
Guangzhou, China  \\
linww@scut.edu.cn}

}

\maketitle

\begin{abstract}
Log anomaly detection is a critical task for system operations and security assurance. However, in networked systems at scale, log data are generated at massive scale while instance-level annotations are prohibitively expensive, posing great difficulties to fine-grained anomaly localization. To address this challenge, we propose LogMILP (Log anomaly localization based on Multi-Instance Learning enhanced by prototypes and Perturbation), a weakly supervised framework that enables both bag-level anomaly detection and instance-level anomaly localization using only bag-level labels. Our method guides the model to pinpoint the critical log entries using prototype-guided structural modeling with counterfactual perturbation consistency regularization, thereby improving localization reliability and interpretability under coarse-grained supervision. Experimental results on three public datasets demonstrate that LogMILP achieves competitive detection performance while yielding significantly more reliable instance-level localization. Our code is open-sourced at \href{https://github.com/YUK1207/LogMILP}{https://github.com/YUK1207/LogMILP}.

\end{abstract}

\begin{IEEEkeywords}
Log Anomaly Detection, Multi-Instance Learning, Weakly-Supervised Learning, Prototype-Guided Mechanism, Perturbation Consistency
\end{IEEEkeywords}

\section{Introduction}
Log data persist as one of the most fundamental sources of operational information in modern networked systems. With the widespread adoption of cloud computing and distributed architectures, log data have grown substantially in scale and semantic complexity, creating difficulties for efficient anomaly detection and precise localization of critical log entries.

Existing log anomaly detection methods generally fit in three categories for label conditions. Supervised methods often achieve strong performance when sufficient annotations are available, but they rely heavily on manual labeling and are therefore difficult to scale to industrial applications \cite{11029770}. Unsupervised methods do not require labeled data, yet they often suffer from high false positive rates when normal and anomalous samples are semantically similar \cite{lyu-etal-2024-towards}. Weakly supervised methods, which use coarse-grained labels, have great practical value but struggle in instance localization and limited interpretability \cite{jiang2023weaklysupervisedanomalydetection}\cite{li2025industrialanomalydetectionlocalization}. 

Considering the nature of log systems and how they are managed, Multi-instance learning (MIL) is well-suited for the scenario: by treating logs in a time window as a \textit{bag}, and each log entry within the window as an \textit{instance}, a detection model can be trained using only bag-level labels\cite{WAQAS2024123893}. This setting closely matches real-world engineering scenarios, \textbf{where the system can only afford window-level alarms rather than precise instance-level annotations}. Although existing MIL-based methods have demonstrated promising potential, they still face two major challenges: 1) instance localization is easily distracted by high-frequency log patterns, and 2) the learned representation does not necessarily reveal causal contribution, impeding the localization of critical entries.

\begin{figure}
    \centering
    \includegraphics[width=1\linewidth]{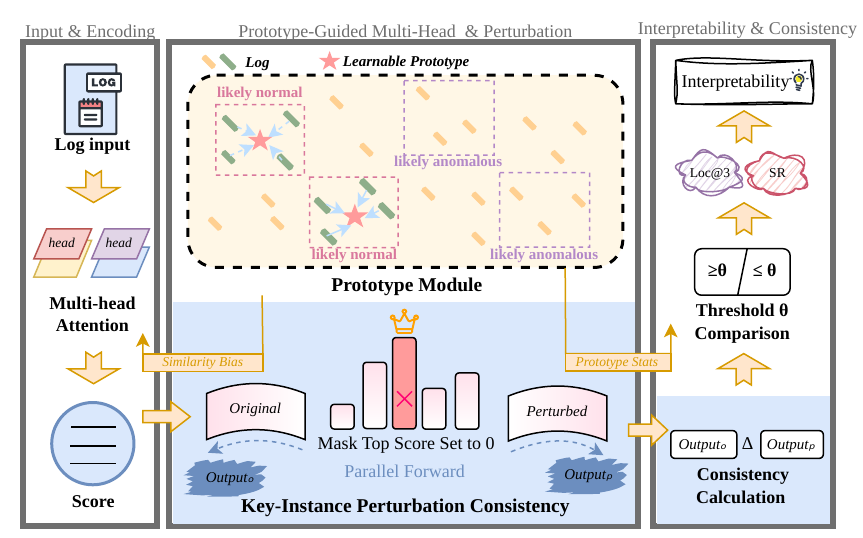}
    \caption{Overall Architecture of LogMILP}
    \label{fig:model}
\end{figure}

To address these issues, we present a Prototype and Perturbation-enhanced Multi-Instance Learning framework (LogMILP) that strengthens the detection model's training with prototype anchors and perturbation sensitivity. Specifically, we use learnable prototype vectors to characterize the distribution of latent patterns and exploit instance-prototype similarity statistics to assist both attention allocation and bag-level prediction. Most importantly, we apply counterfactual perturbation to the key instances identified in each bag to encourage the model to focus on decisive evidence, thereby improving localization reliability and interpretability. The overall architecture of the proposed model is illustrated in Fig.~\ref{fig:model}.

In addition to traditional bag-level evaluation, we also empirically tested our method on instance-level anomaly localization (i.e., finding the critical log entries) using two fine-grained metrics termed Loc@k and Success Rate (SR) \cite{He2025WalkTT}. In summary, our main contributions are as follows:
\begin{itemize}
\item We developed a novel MIL framework tailored for log data mining. To the best of our knowledge, it is the first MIL-based solution to fine-grained log anomaly localization empowered by prototype and perturbation mechanisms.
\item We implemented a unified model architecture that integrates prototype statistical features with multi-head attention, enabling the joint modeling of global pattern distributions and local instance contributions.
\item We introduce a counterfactual perturbation-based training mechanism that effectively mitigates pseudo-localization and improves model interpretability.
\item Extensive experiments on three public datasets, BGL\cite{4273008}\cite{DBLP:journals/corr/abs-2008-06448}, Spirit\cite{4273008}, and ZooKeeper\cite{DBLP:journals/corr/abs-2008-06448} demonstrate that LogMILP achieves clear advantages in both detection performance and localization reliability.
\end{itemize}

\section{Related Work}

\subsection{Log Anomaly Detection}
Early approaches detect anomalies by modeling normal patterns. A representative example is DeepLog \cite{Du17DeepLog}, which employed LSTM to learn the temporal dependencies of log template sequences and regards logs that deviate from the predicted patterns as anomalous. LogAnomaly \cite{ijcai2019p658} further incorporated semantic and statistical features to improve adaptability in complex scenarios. These methods perform well in environments with stable structures and limited template variation, but they usually rely on instance-level labels for supervised training.

With the development of deep representation learning, an increasing number of studies have leveraged contextual semantics to improve detection performance. LogBERT\cite{DBLP:journals/corr/abs-2103-04475} formulates log anomaly detection as a self-supervised learning task and learns robust representations through masked prediction and sequence relationship modeling. LogFormer\cite{cmes.2023.025774} further refines the Transformer architecture to enhance long-range modeling. These approaches are generally more effective for session-level detection. However, their primary focus remains on detection accuracy, with limited attention paid to instance-level localization and interpretability.

\subsection{Weakly Supervised Log Anomaly Detection and MIL}\label{AA}
In practical engineering scenarios, precise instance-level annotations are usually difficult to obtain. This realistic problem has motivated increasing studies on weakly supervised log anomaly detection \cite{jiang2023weaklysupervisedanomalydetection}\cite{li2025industrialanomalydetectionlocalization}. Among these approaches, MIL emerged as a practical match with the common practice of large-scale log systems, where logs are parsed and labeled in batches. In many cases, the system can detect the time window of an anomaly but not the exact point of time. MIL targets at this problem setting by using only bag-level labels, thereby enabling both anomaly detection and instance localization. In recent years, attention-based MIL has been widely applied to weakly supervised video anomaly detection and log analysis. For example, MIDLog \cite{11029770} has demonstrated the practical value of this paradigm in reducing annotation costs.

Nevertheless, prior MIL-based methods have two major limitations. First, instance localization is easily affected by noisy logs, high-frequency templates, or statistical bias. Second, although attention distributions are often treated as a basis for interpretability, high attention does not inherently imply high contribution in MIL. Therefore, how to simultaneously improve localization capability and interpretability under weak supervision remains an open problem.

\subsection{Prototype Learning}
Prototype learning explicitly characterizes representative patterns in the data distribution by introducing a set of learnable prototype vectors in the feature space\cite{luo-etal-2023-prototype}. This paradigm has been widely applied to tasks such as image classification\cite{LIU2026133411}, few-shot learning\cite{ijcai2024p415}, temporal modeling\cite{pmlr-v202-li23d}, and anomaly detection\cite{DBLP:journals/corr/abs-2408-14498}. Compared with deep models that rely solely on implicit representations, prototype-based mechanisms can construct a more structured feature space, thereby improving both discriminative ability and interpretability.

In anomaly detection tasks, prototypes can be used to characterize the centers of dominant patterns and help identify anomalous samples that deviate from the mainstream distribution. In weakly supervised settings, prototype mechanisms provide additional structural constraints in the absence of instance-level labels, thereby enhancing the separability of different instances in the latent space. This is particularly beneficial for log data mining, where normal samples have abundant patterns but anomalies are sparsely distributed.

\subsection{Perturbation Consistency and Interpretability}
In recent years, research in interpretable machine learning has increasingly shown that attention weights or saliency scores do not necessarily reflect the true basis of model decisions\cite{lyu-etal-2024-towards}. On this point, counterfactual perturbation\cite{ijcai2022p103} and consistency regularization\cite{NEURIPS2020_44feb009} have emerged as important mechanisms. The core idea is to delete, mask, or replace the input segments identified by the model as most critical, and then examine whether the output changes as expected.

This idea has been validated in weakly supervised video anomaly detection\cite{s24010058}, natural language processing\cite{qiang-etal-2024-prompt}, and interpretable neural network analysis\cite{ANTAMIS2024128204}. For weakly supervised log anomaly detection, counterfactual perturbation can provide an additional reliability check for instance localization: if removing the instance with the highest attention weight results in almost no change in prediction, the corresponding localization is likely to reflect a spurious correlation rather than true evidence. Motivated by this observation, we propose to incorporate a tailored perturbation consistency regularization into the MIL framework for log anomaly detection, so as to make our model decisions reliable and interpretable.

\section{Methodology}
\subsection{Overview}\label{SCM}
We consider a practical scenario where anomalous event alarms are provided only for time windows, while annotations for individual log entries are absent. We therefore formulate it as a multi-instance learning problem. Accordingly, each time window (or a block of logs) is treated as a \textit{bag} and the log entries within it are regarded as \textit{instances}, with training conducted using only bag-level labels.

LogMILP has three building blocks: instance representation encoding, prototype-guided multi-head attention aggregation, and key-instance perturbation consistency training. The model first applies linear projection and contextual encoding to the input log embeddings to obtain instance-level latent representations. It then leverages learnable prototypes to model representative pattern distributions in the latent space, and uses prototype similarity statistics to assist both attention aggregation and classification. Finally, perturbation samples are constructed based on the key instances identified by the current model, and a consistency constraint is imposed to improve the reliability of instance localization.

\subsection{Problem Statement}
Consider an original log sequence $S = \{x_1, x_2, \dots, x_N\}$ where $x_i \in \mathbb{R}^d$ denotes the input embedding of the $i$-th log entry, the sequence is naturally split with a fixed window size $W$\footnote{For example, logs are parsed and packed every 6 hours.} and stride $s$, yielding a collection of sub-sequences (termed \textbf{bags} in MIL):
\begin{equation}
    B_j = \{x_{(j-1)s+1}, x_{(j-1)s+2}, \dots, x_{(j-1)s+W}\}.
\end{equation}

Each bag is associated with a label $Y_j \in \{0,1\}$. Under the MIL setting,
\begin{equation}
    Y_j = \max_{k=1}^W \{y_{j,k}\},
\end{equation}
where $y_{j,k} = 1$ implies an anomaly event recorded by the $k$-th log instance but is unavailable in the system. During training, \textbf{only the bag-level labels are available}.

\subsection{Instance Encoding}
For each bag $B$, the model first projects the input embeddings into a latent space through a linear transformation: $\mathbf{H} = \mathbf{X}\mathbf{W} + \mathbf{b}$, where $\mathbf{X} \in \mathbb{R}^{W \times d}$ denotes the input sequence, $\mathbf{W} \in \mathbb{R}^{d \times d_h}$ and $\mathbf{b} \in \mathbb{R}^{d_h}$ are learnable parameters, and $d_h$ is the hidden dimension. The resulting representation $\mathbf{H}$ is then fed into a two-layer Transformer Encoder \cite{vaswani2017attention} $\Psi$ to obtain context-enhanced representations: $\mathbf{Z} = \Psi(\mathbf{H})$, where $\mathbf{Z}=\{\mathbf{z}_1,\mathbf{z}_2,\dots,\mathbf{z}_W\}$ and $\mathbf{z}_i \in \mathbb{R}^{d_h}$.

\subsection{Prototype-guided Representation Learning}
To enhance the structured modeling of typical log patterns, we define $N$ learnable prototype vectors, denoted by $P=\{p_1,p_2,\dots,p_N\}$, where $p_j\in\mathbb{R}^{d_h}$. After applying $L_2$ normalization to both the instance representations and the prototypes, the Euclidean distance is computed as $d_{i,j}=\lVert \hat z_i-\hat p_j\rVert_2$, which is then mapped into a similarity score $s_{i,j}=\frac{1}{1+d_{i,j}}$, where $s_{i,j}\in(0,1]$. The maximum prototype similarity for each instance is defined as $m_i=\max_j s_{i,j}$, based on which an anomaly-candidate bias is introduced as $b_i=1-m_i$.

At the bag level, we construct prototype statistical features, including the maximum instance similarity \(M_{bag}=\max_i m_i\), the prototype assignment entropy \(E_{bag}\), and the average prototype activation \(V_{bag}\), which are concatenated as \(F_p=(M_{bag},E_{bag},V_{bag})\). It should be emphasized that \(F_p\) serves as an auxiliary statistical descriptor rather than a direct anomaly score.Finally, the model outputs the bag-level prediction \(\hat{y}\) based on \(F_p\) and \(Z_{cat}\), together with the attention weights \(A\) and intermediate statistics \(\mathcal{E}\).

\subsection{Enforcing Perturbation Consistency in Training}
Relying solely on attention weights can easily lead to pseudo-localization, where instances receive high attention but contribute little causally to the prediction. To address this issue, we introduce a training-time perturbation mechanism: For each bag, we first locate the key index $i^*$ that has the maximum attention score, and then the corresponding embedding is zeroed out to construct a perturbed bag. The prediction (as a probability distribution) before and after perturbation, denoted by $P_{\text{orig}}$ and $P_{\text{pert}}$, are then computed. Therefore, given a positive bag \(\mathcal{B}_{pos}\), the consistency loss is defined as:

\begin{equation}
  \mathcal{L}_{con}=
\frac{1}{|\mathcal{B}_{pos}|}
\sum_{B\in\mathcal{B}_{pos}}
\max\big(0,\ \Delta_c-(P_{orig}-P_{pert})\big),
\end{equation}
where \(\Delta_c\) denotes the consistency margin. \textbf{If the prediction confidence does not drop sufficiently after removing the key instance, a penalty is imposed, thereby encouraging the model to focus on truly critical anomalous evidence.} 

Focal Loss\cite{lin2017focal} is adopted as the primary classification objective, and is jointly optimized with prototype regularization, attention entropy regularization, and consistency loss:
\begin{equation}
    \mathcal{L}_{total}=
\mathcal{L}_{cls}
+\lambda_1\mathcal{L}_{proto}
+\lambda_2\mathcal{L}_{attn}
+\lambda_3\mathcal{L}_{con},
\end{equation}
where \(\lambda_1,\lambda_2,\lambda_3\) denote the corresponding loss weights, the formulation of $\mathcal{L}_{\text{proto}}$, $\mathcal{L}_{\text{attn}}$, and $\mathcal{L}_{\text{attn}}$ are detailed in Algorithm~\ref{alg:train_epoch}. Training is conducted using only bag-level labels, while instance-level labels are not involved in parameter optimization.

\begin{algorithm}[htbp]
\caption{Training Logic with Perturbation Consistency}
\label{alg:train_epoch}
\begin{algorithmic}[1]
\item[\textbf{Input:}] Model $\theta$, loader, optimizer, loss weights $\lambda_p,\lambda_a,\lambda_c$, prototype margins $\Delta_p,\Delta_e$, entropy weight $w_{\text{ent}}$, consistency margin $\Delta_c$, $\epsilon$, $\text{use\_consistency} \in \{\text{T},\text{F}\}$
\item[\textbf{Output:}] Updated model parameters $\theta$

\FOR{each mini-batch $(X,Y)$}
  \STATE $(P_{\text{orig}}, \mathcal{A}, \mathcal{E}) \gets \text{model}(X)$
  \STATE $\mathcal{I}_{\text{pos}} \gets \{\, b \mid Y^{(b)} = 1 \,\},\quad
         \mathcal{I}_{\text{neg}} \gets \{\, b \mid Y^{(b)} = 0 \,\}$
  \STATE $\mathcal{L}_{\text{cls}} \gets \text{FocalLoss}(P_{\text{orig}}, Y)$

  \STATE $\mathcal{L}_{\text{proto}}^{\text{pos}} \gets
  \operatorname{mean}_{b \in \mathcal{I}_{\text{pos}}}
  \bigl(\max(0,\Delta_p - M_{bag}^{(b)})\bigr)$

  \STATE $\mathcal{L}_{\text{proto}}^{\text{neg}} \gets
  \operatorname{mean}_{b \in \mathcal{I}_{\text{neg}}}
  \bigl(\max(0,\Delta_e - E_{bag}^{(b)})\bigr)$

  \STATE $\mathcal{L}_{\text{proto}} \gets
  \mathcal{L}_{\text{proto}}^{\text{pos}} + w_{\text{ent}}\,\mathcal{L}_{\text{proto}}^{\text{neg}}$

  \STATE $\mathcal{L}_{\text{attn}} \gets
  \operatorname{mean}\!\Bigl(-\sum_t \mathcal{A}_{:,:,t}\log(\mathcal{A}_{:,:,t}+\epsilon)\big/\log(W)\Bigr)$
  
    \STATE $k_b^* \gets \arg\min_k \operatorname{Entropy}(\mathcal{A}[b,k,:]),\ \forall b$
    \STATE $i_b^* \gets \arg\max_t \mathcal{A}[b,k_b^*,t],\ \forall b$
    \STATE $\tilde{X} \gets X$
    \STATE $\tilde{X}[b,i_b^*,:] \gets 0,\ \forall b$
    \STATE $(P_{\text{pert}}, -, -) \gets \text{model}(\tilde{X})$
    \STATE $\mathcal{L}_{\text{con}} \gets
    \operatorname{mean}_{b \in \mathcal{I}_{\text{pos}}}
    \bigl(\max(0,\Delta_c-(P_{\text{orig}}^{(b)}-P_{\text{pert}}^{(b)}))\bigr)$

  \STATE $\mathcal{L}_{\text{total}} \gets
  \mathcal{L}_{\text{cls}} + \lambda_p \mathcal{L}_{\text{proto}} + \lambda_a \mathcal{L}_{\text{attn}} + \lambda_c \mathcal{L}_{\text{con}}$
  \STATE Compute gradients $\nabla_\theta\mathcal{L}_{\text{cls}}$ and do back-propagation
\ENDFOR
\RETURN $\theta$
\end{algorithmic}
\end{algorithm}

\subsection{Localizing Instance-level Anomalies}
\label{sec:metrics}
For each bag $B$ of logs labeled positive, we examine the backbone model's attention head with the minimum attention entropy, and then \textbf{identify the anomaly candidates as the top-k instances with the highest attention weights} in that head, denoted as \(S^{top}_B\). 

Empirically, we evaluate instance-level anomaly localization accuracy by two metrics:
\subsubsection{Loc@k}
Let the set of ground-truth anomalous instances be \(S_B^{a}\), we define the localization hit rate $Loc@k$ by:
\begin{equation}
    Loc@k=
\frac{\sum_{B\in\mathcal{B}_{pos}} |S^{top}_B\cap S^{a}_B|}
{\sum_{B\in\mathcal{B}_{pos}} \min(k,\ |S^{a}_B|)},
\end{equation}

\subsubsection{Success Rate}
Again, we use the perturbation mechanism to test whether the localization is reliable. For each positive bag, indexed by \(b\), we compare the model-predicted bag-level anomaly probabilities before and after removing the key instance, denoted by \(P_{\text{orig}}^{(b)}\) and \(P_{\text{pert}}^{(b)}\), respectively. On this basis, we define the Success Rate (SR) as
\begin{equation}
\mathrm{SR}
=
\frac{1}{|\mathcal{B}_{pos}|}
\sum_{b=1}^{|\mathcal{B}_{pos}|}
\mathbbm{1}\!\left(P_{\text{orig}}^{(b)} - P_{\text{pert}}^{(b)} > \delta_{sr}\right),
\end{equation}
where \(\mathbbm{1}(\cdot)\) is the indicator function. A higher SR indicates that the model relies more on truly decision-critical instances rather than incidental correlated patterns.

\begin{table*}[!tb]
\centering
\caption{Comparison of bag-level anomaly detection performance across datasets}
\label{tab:main_results}
\TableOneStyle
\begin{adjustbox}{max width=\textwidth}
\begin{tabular}{l|cccc|cccc|cccc}
\toprule
\multirow{2}{*}{\textbf{Model}} & \multicolumn{4}{c|}{\textbf{BGL}} & \multicolumn{4}{c|}{\textbf{Spirit}} & \multicolumn{4}{c}{\textbf{ZooKeeper}} \\ \cmidrule(lr){2-5} \cmidrule(lr){6-9} \cmidrule(lr){10-13}
 & AUC & Prec. & Rec. & F1 & AUC & Prec. & Rec. & F1 & AUC & Prec. & Rec. & F1 \\ \midrule
DeepLog\cite{Du17DeepLog} & 0.9233 & 0.7004 & 0.8409 & 0.7643 & 0.8929 & 0.7901 & 0.9191 & 0.8497 & 0.9991 & 0.9999 & 0.9898 & 0.9948 \\
LogAnomaly\cite{ijcai2019p658} & 0.9438 & 0.7512 & 0.7109 & 0.7305 & 0.8067 & 0.6838 & 0.9204 & 0.7846 & 0.9993 & 0.9999 & 0.9882 & 0.9940 \\
LogBERT\cite{DBLP:journals/corr/abs-2103-04475} & 0.9408 & 0.8894 & 0.7594 & 0.8193 & 0.9633 & 0.9439 & 0.9008 & 0.9218 & 0.9994 & 1.0000 & 0.9812 & 0.9905 \\
LogFormer\cite{cmes.2023.025774} & 0.9216 & 0.7085 & 0.7154 & 0.7119 & 0.5035 & 0.5855 & 0.7507 & 0.6579 & 0.9989 & 1.0000 & 0.9923 & 0.9961 \\
MIDLog\cite{11029770} & 0.9752 & 0.9494 & 0.8254 & 0.8830 & 0.9668 & 0.9195 & 0.9243 & 0.9219 & 0.9870 & 1.0000 & 0.9684 & 0.9840 \\ \midrule
\textbf{OURS} & 0.9464 & 0.9264 & 0.9421 & \textbf{0.9342} & 0.9652 & 0.9194 & 0.9404 & \textbf{0.9295} & 0.9964 & 0.9964 & 0.9970 & \textbf{0.9967} \\
(std) & $\pm$0.0158 & $\pm$0.0118 & $\pm$0.0096 & $\boldsymbol{\pm}{0.0101}$ & $\pm$0.0068 & $\pm$0.0230 & $\pm$0.0183 & $\boldsymbol{\pm}{0.0089}$ & $\pm$0.0026 & $\pm$0.0026 & $\pm$0.0015 & $\boldsymbol{\pm}{0.0011}$ \\
\bottomrule
\end{tabular}
\end{adjustbox}
\end{table*}

\begin{figure*}[!htb]
    \centering
    \includegraphics[width=0.98\textwidth]{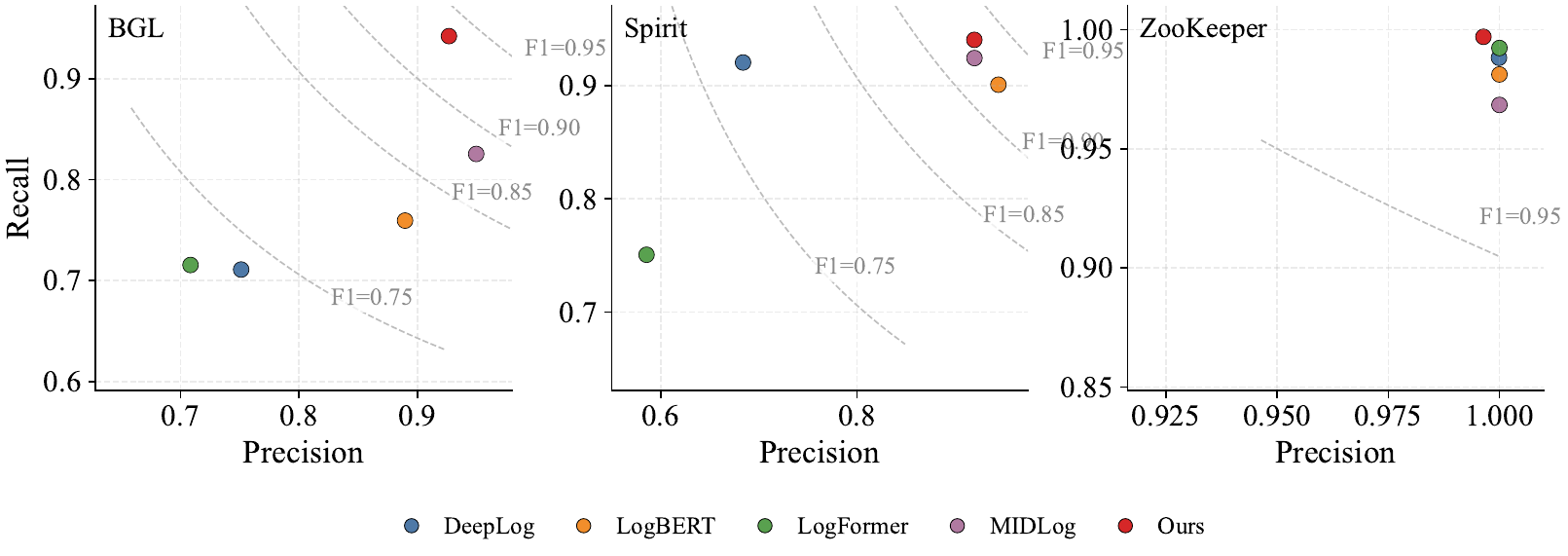}
    \caption{Geometric distribution of baseline models in the precision-recall space}
    \label{fig:pr_geometry}
\end{figure*}

\section{Experimental evaluation}
\subsection{Experimental Setup}
\subsubsection{Datasets}
We evaluated the proposed method on three public datasets for log anomaly detection: BGL\cite{4273008,DBLP:journals/corr/abs-2008-06448}, Spirit\cite{4273008}, and ZooKeeper\cite{DBLP:journals/corr/abs-2008-06448}. All raw logs were processed through a unified pre-processing pipeline and subsequently organized into multi-instance bags according to their temporal or logical structure. Specifically, BGL and ZooKeeper logs were bagged using sliding time windows, whereas Spirit used non-overlapping blocks that are further aggregated into bags with a fixed number of instances. 




\subsubsection{Baselines}
We drew comparison with DeepLog\cite{Du17DeepLog}, LogAnomaly\cite{ijcai2019p658}, LogBERT\cite{DBLP:journals/corr/abs-2103-04475}, LogFormer\cite{cmes.2023.025774}, and MIDLog\cite{11029770}. DeepLog and LogAnomaly represent classical sequence modeling approaches, while LogBERT and LogFormer represent advanced methods based on pretrained semantics and Transformer architectures. MIDLog serves as the weakly supervised MIL baseline most closely related to our method.

All baseline models are evaluated under a unified data pre-processing pipeline and bag-level evaluation protocol. It should be noted that the original designs of LogBERT and LogFormer are not directly intended for instance-level localization or perturbation-consistency evaluation. In this work, we introduce only offline instance scoring and perturbation-based evaluation adaptations to compute Loc@3 and SR, without modifying their core modeling logic for the bag-level detection task. Accordingly, these results are interpreted as supplementary instance-level comparisons rather than evidence that such capabilities are natively supported by the original models.



\subsubsection{Evaluation Protocols}
All experiments were conducted on a Linux platform equipped with an Intel(R) Xeon(R) Platinum 8470Q CPU and an NVIDIA GeForce RTX 5090 GPU. All experiments were repeated with three random seeds. To address the class imbalance issue in weakly supervised settings, a \texttt{WeightedRandomSampler} was employed during training.

For any method applicable, we include both bag-level detection metrics and instance-level reliability metrics. For the bag-level detection task, F1 score is used as the primary metric. Given the output probability $P$, the optimal threshold $\tau$ is first selected on the validation set and then applied to the test set to compute the final Precision, Recall, and F1 scores.

At the instance level, we use Loc@3 and SR, defined in Sec. \ref{sec:metrics}, to measure localization accuracy and causal reliability, respectively. During training, the model is optimized using only bag-level labels. The computation of Loc@3 and SR is performed only during testing, using the instance-level ground truth already available in the datasets for offline evaluation, and does not participate in training or threshold selection.

\subsection{Main Results}
We first report bag-level results in conventional metrics, and then demonstrate the effectiveness of LogMILP with instance-level metrics.
\subsubsection{Performance on Bag-level Anomaly Detection}
Overall, LogMILP achieved the best F1 scores (0.9342, 0.9295 and 0.9967) across all three datasets (Table~\ref{tab:main_results}). Especially, a significant boost in recall was observed on BGL ($>$10\% gap over the 2nd best). To further compare the operating characteristics of different methods in the precision-recall space, Fig.~\ref{fig:pr_geometry} visualizes the results with iso-F1 curves.



In addition, we observe that sequence matching-based methods, such as DeepLog and LogAnomaly, suffer a substantial performance degradation under coarse-grained weak supervision, suggesting that they are highly dependent on precise instance-level annotations. We note that the performance of all methods on the ZooKeeper dataset is close to the ceiling, indicating that bag-level supervision could be sufficient in such systems. Nonetheless, it does not necessarily mean that existing methods can also work well for instance-level anomaly localization on the same condition.

\begin{table*}[!tb]
\centering
\caption{Comparison of instance-level anomaly localization in terms of Loc@3 and SR}
\label{tab:sr_loc_comparison}
\TableOneStyle
\begin{adjustbox}{max width=\textwidth}
\begin{tabular}{lcccccc}
\toprule
\multirow{2}{*}{\textbf{Model}}
& \multicolumn{2}{c}{\textbf{BGL}}
& \multicolumn{2}{c}{\textbf{Spirit}}
& \multicolumn{2}{c}{\textbf{ZooKeeper}} \\
\cmidrule(lr){2-3} \cmidrule(lr){4-5} \cmidrule(lr){6-7}
& Loc@3 & SR & Loc@3 & SR & Loc@3 & SR \\
\midrule
*LogBERT   & 0.3794 & 0.7755 & 0.6569 & 0.5953 & 0.8261 & 0.8696 \\
*LogFormer & 0.3185 & 0.9040 & 0.1912 & 0.9387 & 0.8604 & \textbf{0.9979} \\
\midrule
\textbf{OURS} & $\bm{0.3488} \pm 0.0312$ & $\bm{0.9730} \pm 0.0147$ & $\bm{0.7786} \pm 0.0255$ & $\bm{0.9658} \pm 0.0148$ & $\bm{0.8917} \pm 0.0346$ & ${0.9962} \pm 0.0066$ \\
\midrule
\multicolumn{7}{l}{\textit{* indicates an offline instance-level evaluation adaptation for Loc@3/SR without changing the core bag-level logic of the original model.}} \\
\bottomrule
\end{tabular}
\end{adjustbox}
\end{table*}


\subsubsection{Performance on Instance-level Anomaly Localization}
In this part, we compared different methods in terms of instance localization quality and reliability. It should be noted that LogBERT and LogFormer were partially adapted in this section for the comparison of Loc@3 and SR, which follows the attention score-based approaches. LogMILP achieved high SR across all three datasets, while logBERT struggled to offer reliable decisions at instance-level. Results also show that LogMILP outperformed the baselines in Loc@3 on the Spirit dataset by a large margin, offering strong insight for locating the critical log entries.

\subsection{Ablation Study}

To verify the contribution of our perturbation consistency mechanism, we compare the full model with a variant without consistency loss while keeping all other components unchanged. The results are reported in Table~\ref{tab:consistency_ablation_vertical}. It can be observed that consistency regularization significantly improves localization reliability, particularly in terms of the SR metric. This verifies that counterfactual perturbation effectively forces the model to learn the sources of anomaly without relying on instance-level labels.

\begin{table}[t]
\centering
\caption{Consistency loss ablation results. $\Delta$ denotes the performance gap between the full model and the ablated version (Full - w/o consistency loss).}
\label{tab:consistency_ablation_vertical}
\small
\begin{tabular}{llcc}
\toprule
Dataset & Metric & w/o consistency loss & $\Delta$ (Full - w/o) \\
\midrule
\multirow{5}{*}{BGL} 
& Precision & $0.9314 \pm 0.0058$ & $-0.0050$ \\
& Recall    & $0.8682 \pm 0.0117$ & $+0.0739$ \\
& F1        & $0.8987 \pm 0.0039$ & $+0.0355$ \\
& Loc@3     & $0.2485 \pm 0.0422$ & $+0.1003$ \\
& SR        & $0.0459 \pm 0.0249$ & $+0.9271$ \\
\midrule
\multirow{5}{*}{Spirit} 
& Precision & $0.8984 \pm 0.0173$ & $+0.0210$ \\
& Recall    & $0.9235 \pm 0.0183$ & $+0.0169$ \\
& F1        & $0.9105 \pm 0.0024$ & $+0.0190$ \\
& Loc@3     & $0.6039 \pm 0.0956$ & $+0.1747$ \\
& SR        & $0.0104 \pm 0.0046$ & $+0.9554$ \\
\midrule
\multirow{5}{*}{ZooKeeper} 
& Precision & $0.9964 \pm 0.0015$ & $0.0000$ \\
& Recall    & $0.9630 \pm 0.0190$ & $+0.0340$ \\
& F1        & $0.9794 \pm 0.0098$ & $+0.0173$ \\
& Loc@3     & $0.8722 \pm 0.0720$ & $+0.0195$ \\
& SR        & $0.0133 \pm 0.0121$ & $+0.9829$ \\
\bottomrule
\end{tabular}
\end{table}

\section{Conclusion}
Log anomaly detection is a critical problem in AIOps and cybersecurity. In large-scale industrial scenarios, fine-grained instance-level annotations are often difficult to obtain, making weakly supervised MIL a more practical modeling paradigm. Existing methods mainly focus on bag-level detection, with relatively limited systematic attention paid to instance localization capability and the reliability of its interpretation. In this paper, we propose LogMILP, which unifies learnable prototype guidance, multi-head attention aggregation, and key-instance perturbation consistency training within a single MIL framework. Using only bag-level labels, the proposed method simultaneously improves detection performance, localization capability, and localization reliability. Experimental results on three public datasets, BGL, Spirit, and ZooKeeper, demonstrate that the proposed method is highly competitive on bag-level metrics while showing clear advantages on instance-level metrics such as Loc@3 and SR.

Our work offers a practically viable solution but still has limitations such as untested robustness to noisy data. Future plan of research will include incremental prototype updating for online scenarios, deeper integration with large-scale pretrained log representations as well as cross-domain generalization in more complex industrial log streams.

\bibliographystyle{ieeetr}
\bibliography{ref}

@article{WAQAS2024123893,
title = {Exploring Multiple Instance Learning (MIL): A brief survey},
journal = {Expert Systems with Applications},
volume = {250},
pages = {123893},
year = {2024},
issn = {0957-4174},
doi = {https://doi.org/10.1016/j.eswa.2024.123893},
url = {https://www.sciencedirect.com/science/article/pii/S0957417424007590},
author = {Muhammad Waqas and Syed Umaid Ahmed and Muhammad Atif Tahir and Jia Wu and Rizwan Qureshi}
}

@article{He2025WalkTT,
  title={Walk the Talk: Is Your Log-based Software Reliability Maintenance System Really Reliable?},
  author={Minghua He and Tong Jia and Chiming Duan and Pei Xiao and Lingzhe Zhang and Kangjin Wang and Yifan Wu and Ying Li and Gang Huang},
  journal={2025 40th IEEE/ACM International Conference on Automated Software Engineering (ASE)},
  year={2025},
  pages={3784-3788},
  url={https://api.semanticscholar.org/CorpusID:281675075}
}

@article{LIU2026133411,
title = {Confident classification via template representation learning},
journal = {Neurocomputing},
volume = {682},
pages = {133411},
year = {2026},
issn = {0925-2312},
doi = {https://doi.org/10.1016/j.neucom.2026.133411},
url = {https://www.sciencedirect.com/science/article/pii/S0925231226008088},
author = {Yangyang Liu and Fei Yin and Cheng-Lin Liu}
}

@InProceedings{pmlr-v202-li23d,
  title = 	 {Prototype-oriented unsupervised anomaly detection for multivariate time series},
  author =       {Li, Yuxin and Chen, Wenchao and Chen, Bo and Wang, Dongsheng and Tian, Long and Zhou, Mingyuan},
  booktitle = 	 {Proceedings of the 40th International Conference on Machine Learning},
  pages = 	 {19407--19424},
  year = 	 {2023},
  editor = 	 {Krause, Andreas and Brunskill, Emma and Cho, Kyunghyun and Engelhardt, Barbara and Sabato, Sivan and Scarlett, Jonathan},
  volume = 	 {202},
  series = 	 {Proceedings of Machine Learning Research},
  month = 	 {23--29 Jul},
  publisher =    {PMLR},
  url = 	 {https://proceedings.mlr.press/v202/li23d.html}
}

@article{DBLP:journals/corr/abs-2408-14498,
  publtype={informal},
  author={Zhijin Dong and Hongzhi Liu and Boyuan Ren and Weimin Xiong and Zhonghai Wu},
  title={Reconstruction-based Multi-Normal Prototypes Learning for Weakly Supervised Anomaly Detection},
  year={2024},
  cdate={1704067200000},
  journal={CoRR},
  volume={abs/2408.14498},
  url={https://doi.org/10.48550/arXiv.2408.14498}
}

@inproceedings{ijcai2022p103,
  title     = {Counterfactual Interpolation Augmentation (CIA): A Unified Approach to Enhance Fairness and Explainability of DNN},
  author    = {Qiang, Yao and Li, Chengyin and Brocanelli, Marco and Zhu, Dongxiao},
  booktitle = {Proceedings of the Thirty-First International Joint Conference on
               Artificial Intelligence, {IJCAI-22}},
  publisher = {International Joint Conferences on Artificial Intelligence Organization},
  editor    = {Lud De Raedt},
  pages     = {732--739},
  year      = {2022},
  month     = {7},
  note      = {Main Track},
  doi       = {10.24963/ijcai.2022/103},
  url       = {https://doi.org/10.24963/ijcai.2022/103},
}

@inproceedings{NEURIPS2020_44feb009,
 author = {Xie, Qizhe and Dai, Zihang and Hovy, Eduard and Luong, Thang and Le, Quoc},
 booktitle = {Advances in Neural Information Processing Systems},
 editor = {H. Larochelle and M. Ranzato and R. Hadsell and M.F. Balcan and H. Lin},
 pages = {6256--6268},
 publisher = {Curran Associates, Inc.},
 title = {Unsupervised Data Augmentation for Consistency Training},
 url = {https://proceedings.neurips.cc/paper_files/paper/2020/file/44feb0096faa8326192570788b38c1d1-Paper.pdf},
 volume = {33},
 year = {2020}
}

@Article{s24010058,
AUTHOR = {Lee, Junyeop and Koo, Hyunbon and Kim, Seongjun and Ko, Hanseok},
TITLE = {Cognitive Refined Augmentation for Video Anomaly Detection in Weak Supervision},
JOURNAL = {Sensors},
VOLUME = {24},
YEAR = {2024},
NUMBER = {1},
ARTICLE-NUMBER = {58},
URL = {https://www.mdpi.com/1424-8220/24/1/58},
PubMedID = {38202920},
ISSN = {1424-8220},
DOI = {10.3390/s24010058}
}

@article{ANTAMIS2024128204,
title = {Interpretability of deep neural networks: A review of methods, classification and hardware},
journal = {Neurocomputing},
volume = {601},
pages = {128204},
year = {2024},
issn = {0925-2312},
doi = {https://doi.org/10.1016/j.neucom.2024.128204},
url = {https://www.sciencedirect.com/science/article/pii/S0925231224009755},
author = {Thanasis Antamis and Anastasis Drosou and Thanasis Vafeiadis and Alexandros Nizamis and Dimosthenis Ioannidis and Dimitrios Tzovaras}
}

@INPROCEEDINGS{4273008,
  author={Oliner, Adam and Stearley, Jon},
  booktitle={37th Annual IEEE/IFIP International Conference on Dependable Systems and Networks (DSN'07)}, 
  title={What Supercomputers Say: A Study of Five System Logs}, 
  year={2007},
  volume={},
  number={},
  pages={575-584},
  doi={10.1109/DSN.2007.103}}

@article{DBLP:journals/corr/abs-2008-06448,
  author       = {Shilin He and
                  Jieming Zhu and
                  Pinjia He and
                  Michael R. Lyu},
  title        = {Loghub: {A} Large Collection of System Log Datasets towards Automated
                  Log Analytics},
  journal      = {CoRR},
  volume       = {abs/2008.06448},
  year         = {2020},
  url          = {https://arxiv.org/abs/2008.06448},
  eprinttype   = {arXiv},
  eprint       = {2008.06448},
  bibsource    = {dblp computer science bibliography, https://dblp.org}
}

@inproceedings{
Du17DeepLog,
Author = {Min Du and Feifei Li and Guineng Zheng and Vivek Srikumar},
Title = {{DeepLog:} Anomaly Detection and Diagnosis from System Logs through Deep Learning},
Booktitle = {ACM Conference on Computer and Communications Security (CCS)},
Year = {2017}
}

@inproceedings{ijcai2019p658,
  title     = {LogAnomaly: Unsupervised Detection of Sequential and Quantitative Anomalies in Unstructured Logs},
  author    = {Meng, Weibin and Liu, Ying and Zhu, Yichen and Zhang, Shenglin and Pei, Dan and Liu, Yuqing and Chen, Yihao and Zhang, Ruizhi and Tao, Shimin and Sun, Pei and Zhou, Rong},
  booktitle = {Proceedings of the Twenty-Eighth International Joint Conference on
               Artificial Intelligence, {IJCAI-19}},
  publisher = {International Joint Conferences on Artificial Intelligence Organization},
  pages     = {4739--4745},
  year      = {2019},
  month     = {7},
  doi       = {10.24963/ijcai.2019/658},
  url       = {https://doi.org/10.24963/ijcai.2019/658},
}

@article{DBLP:journals/corr/abs-2103-04475,
  author       = {Haixuan Guo and
                  Shuhan Yuan and
                  Xintao Wu},
  title        = {LogBERT: Log Anomaly Detection via {BERT}},
  journal      = {CoRR},
  volume       = {abs/2103.04475},
  year         = {2021},
  url          = {https://arxiv.org/abs/2103.04475},
  eprinttype   = {arXiv},
  eprint       = {2103.04475},
  bibsource    = {dblp computer science bibliography, https://dblp.org}
}

@Article{cmes.2023.025774,
  author   = {Feilu Hang and Wei Guo and Hexiong Chen and Linjiang Xie and Chenghao Zhou and Yao Liu},
  title    = {Logformer: Cascaded Transformer for System Log Anomaly Detection},
  journal  = {Computer Modeling in Engineering \& Sciences},
  volume   = {136},
  year     = {2023},
  number   = {1},
  pages    = {517--529},
  url      = {http://www.techscience.com/CMES/v136n1/51217},
  issn     = {1526-1506},
  doi      = {10.32604/cmes.2023.025774}
}

@INPROCEEDINGS{11029770,
  author={He, Minghua and Jia, Tong and Duan, Chiming and Cai, Huaqian and Li, Ying and Huang, Gang},
  booktitle={2025 IEEE/ACM 47th International Conference on Software Engineering (ICSE)}, 
  title={Weakly-Supervised Log-Based Anomaly Detection with Inexact Labels via Multi-Instance Learning}, 
  year={2025},
  pages={2918-2930},
  doi={10.1109/ICSE55347.2025.00189}}

@misc{jiang2023weaklysupervisedanomalydetection,
      title={Weakly Supervised Anomaly Detection: A Survey}, 
      author={Minqi Jiang and Chaochuan Hou and Ao Zheng and Xiyang Hu and Songqiao Han and Hailiang Huang and Xiangnan He and Philip S. Yu and Yue Zhao},
      year={2023},
      eprint={2302.04549},
      archivePrefix={arXiv},
      primaryClass={cs.LG},
      url={https://arxiv.org/abs/2302.04549}, 
}

@misc{li2025industrialanomalydetectionlocalization,
      title={Industrial Anomaly Detection and Localization Using Weakly-Supervised Residual Transformers}, 
      author={Hanxi Li and Jingqi Wu and Deyin Liu and Lin Wu and Hao Chen and Mingwen Wang and Chunhua Shen},
      year={2025},
      eprint={2306.03492},
      archivePrefix={arXiv},
      primaryClass={cs.CV},
      url={https://arxiv.org/abs/2306.03492}, 
}

@inproceedings{vaswani2017attention,
  title     = {Attention Is All You Need},
  author    = {Vaswani, Ashish and Shazeer, Noam and Parmar, Niki and Uszkoreit, Jakob and Jones, Llion and Gomez, Aidan N. and Kaiser, {\L}ukasz and Polosukhin, Illia},
  booktitle = {Advances in Neural Information Processing Systems},
  volume    = {30},
  year      = {2017},
  url       = {https://arxiv.org/abs/1706.03762}
}

@inproceedings{lin2017focal,
  title     = {Focal Loss for Dense Object Detection},
  author    = {Lin, Tsung-Yi and Goyal, Priya and Girshick, Ross and He, Kaiming and Doll{\'a}r, Piotr},
  booktitle = {Proceedings of the IEEE International Conference on Computer Vision (ICCV)},
  pages     = {2980--2988},
  year      = {2017},
  doi       = {10.1109/ICCV.2017.324},
  url       = {https://arxiv.org/abs/1708.02002}
}

@inproceedings{ijcai2024p415,
  title     = {With a Little Help from Language: Semantic Enhanced Visual Prototype Framework for Few-Shot Learning},
  author    = {Cai, Hecheng and Liu, Yang and Huang, Shudong and Lv, Jiancheng},
  booktitle = {Proceedings of the Thirty-Third International Joint Conference on
               Artificial Intelligence, {IJCAI-24}},
  publisher = {International Joint Conferences on Artificial Intelligence Organization},
  editor    = {Kate Larson},
  pages     = {3751--3759},
  year      = {2024},
  month     = {8},
  note      = {Main Track},
  doi       = {10.24963/ijcai.2024/415},
  url       = {https://doi.org/10.24963/ijcai.2024/415},
}

@inproceedings{qiang-etal-2024-prompt,
    title = "Prompt Perturbation Consistency Learning for Robust Language Models",
    author = "Qiang, Yao  and
      Nandi, Subhrangshu  and
      Mehrabi, Ninareh  and
      Ver Steeg, Greg  and
      Kumar, Anoop  and
      Rumshisky, Anna  and
      Galstyan, Aram",
    editor = "Graham, Yvette  and
      Purver, Matthew",
    booktitle = "Findings of the Association for Computational Linguistics: EACL 2024",
    month = mar,
    year = "2024",
    address = "St. Julian{'}s, Malta",
    publisher = "Association for Computational Linguistics",
    url = "https://aclanthology.org/2024.findings-eacl.91/",
    doi = "10.18653/v1/2024.findings-eacl.91",
    pages = "1357--1370"
}

@inproceedings{luo-etal-2023-prototype,
    title = "Prototype-Based Interpretability for Legal Citation Prediction",
    author = "Luo, Chu Fei  and
      Bhambhoria, Rohan  and
      Dahan, Samuel  and
      Zhu, Xiaodan",
    editor = "Rogers, Anna  and
      Boyd-Graber, Jordan  and
      Okazaki, Naoaki",
    booktitle = "Findings of the Association for Computational Linguistics: ACL 2023",
    month = jul,
    year = "2023",
    address = "Toronto, Canada",
    publisher = "Association for Computational Linguistics",
    url = "https://aclanthology.org/2023.findings-acl.301/",
    doi = "10.18653/v1/2023.findings-acl.301",
    pages = "4883--4898"
}

@article{lyu-etal-2024-towards,
    title = "Towards Faithful Model Explanation in {NLP}: A Survey",
    author = "Lyu, Qing  and
      Apidianaki, Marianna  and
      Callison-Burch, Chris",
    journal = "Computational Linguistics",
    volume = "50",
    number = "2",
    month = jun,
    year = "2024",
    address = "Cambridge, MA",
    publisher = "MIT Press",
    url = "https://aclanthology.org/2024.cl-2.6/",
    doi = "10.1162/coli_a_00511",
    pages = "657--723"
}

\end{document}